\documentclass[10pt,twocolumn,letterpaper]{article}

\usepackage{iccv}              
\iccvfinalcopy 

\usepackage{graphicx}
\usepackage{amssymb}
\usepackage{booktabs}

\usepackage{algorithm}
\usepackage{algorithmic}
\usepackage{times}
\usepackage{epsfig}
\usepackage{graphicx}
\usepackage{amssymb}
\usepackage{pifont}
\usepackage{multirow}
\usepackage{xcolor}
\usepackage{bm}
\usepackage{booktabs}
\usepackage{color, colortbl}
\usepackage{lipsum}
\usepackage{subcaption}
\usepackage{here}

\usepackage{amsmath,amsfonts,bm}

\def\eqref#1{equation~\ref{#1}}

\def\1{\bm{1}}

\DeclareMathAlphabet{\mathsfit}{\encodingdefault}{\sfdefault}{m}{sl}
\SetMathAlphabet{\mathsfit}{bold}{\encodingdefault}{\sfdefault}{bx}{n}

\usepackage[normalem]{ulem}

\newcommand\blfootnote[1]{%
  \begingroup
  \renewcommand\thefootnote{}\footnote{#1}%
  \addtocounter{footnote}{-1}%
  \endgroup
}

\usepackage[pagebackref,breaklinks,colorlinks]{hyperref}

\usepackage[capitalize]{cleveref}
\crefname{section}{Sec.}{Secs.}
\Crefname{section}{Section}{Sections}
\Crefname{table}{Table}{Tables}
\crefname{table}{Tab.}{Tabs.}

\def\iccvPaperID{3912} 

\ificcvfinal\fi

\begin{document}

\title{Boosting Novel Category Discovery Over Domains \\with Soft Contrastive Learning and All in One Classifier}

\author{\small Zelin Zang$^\ast$\\
   \small Westlake University\\
   {\tt\small zangzelin@westlake.edu.cn}
   \and
   \small Lei Shang$^\ast$\\
   \small Alibaba Group\\
   {\tt\small sl172005@alibaba-inc.com}
   \and
   \small Senqiao Yang$^\ast$\\
   \small Westlake University\\
   {\tt\small yangsenqiao.ai@gmail.com}
   \and
   \small Fei Wang, Baigui Sun, Xuansong Xie\\
   \small Alibaba Group\\
   {\tt\small \{steven.wf\}\{baigui.sbg\}\{xingtong.xxs\}@alibaba-inc.com}
   \and
   \small Stan Z. Li$^\dagger$\\
   \small Westlake University\\
   {\tt\small Stan.ZQ.Li@westlake.edu.cn}
}

\maketitle

\newcommand{\ourss}{SAN }
\newcommand{\ours}{SAN}
\renewcommand{\thefootnote}{\fnsymbol{footnote}}

\renewcommand{\thefootnote}{\arabic{footnote}}
\begin{abstract}
   \begin{NoHyper}
      \blfootnote{The $^\ast$ indicates equal contribution, $^\dagger$ indicates corresponding author.}
   \end{NoHyper}
   \vspace{-0mm}
Unsupervised domain adaptation (UDA) has proven to be highly effective in transferring knowledge from a label-rich source domain to a label-scarce target domain. However, the presence of additional novel categories in the target domain has led to the development of open-set domain adaptation (ODA) and universal domain adaptation (UNDA). Existing ODA and UNDA methods treat all novel categories as a single, unified unknown class and attempt to detect it during training. However, we found that domain variance can lead to more significant view-noise in unsupervised data augmentation, which affects the effectiveness of contrastive learning (CL) and causes the model to be overconfident in novel category discovery. To address these issues, a framework named Soft-contrastive All-in-one Network (SAN) is proposed for ODA and UNDA tasks. SAN includes a novel data-augmentation-based soft contrastive learning (SCL) loss to fine-tune the backbone for feature transfer and a more human-intuitive classifier to improve new class discovery capability. The SCL loss weakens the adverse effects of the data augmentation view-noise problem which is amplified in domain transfer tasks. The All-in-One (AIO) classifier overcomes the overconfidence problem of current mainstream closed-set and open-set classifiers. Visualization and ablation experiments demonstrate the effectiveness of the proposed innovations. Furthermore, extensive experiment results on ODA and UNDA show that SAN outperforms existing state-of-the-art methods.

\end{abstract}

\section{Introduction}

Domain adaptation (DA) is a technique that transfers knowledge from label-rich training domains to new domains where labels are scarce~\cite{ben2010theory}. It addresses the problem of generalization of deep neural networks in new domains. Traditional unsupervised domain adaptation (UDA) assumes that the source domain and the target domain completely share the sets of categories, i.e., closed-set DA. However, this assumption does not often hold in practice. There are several possible situations, such as the target domain containing types absent in the source (unknown categories), i.e., open-set DA (ODA)~\cite{busto2017open,saito2018open}, the source domain including classes absent in the target (source-private categories), i.e., partial DA (PDA)~\cite{cao2018partial}, or a mixture of ODA and PDA, called open-partial DA (OPDA). Although many approaches have been tailored to a specific setting, we cannot know the category shift in advance, which is an actual difficulty. To account for the uncertainty about the category shift, the task of universal domain adaptation (UNDA) is proposed~\cite{UDA_2019_CVPR, saito2020universal}. The assumption is that the label distributions of labeled and unlabeled data can differ, but we do not know the difference in advance. UNDA is a uniform and practical setting since estimating the label distributions of unlabeled data is very hard in real applications.

\begin{figure*}
  \centering
  \includegraphics[width=0.99\linewidth]{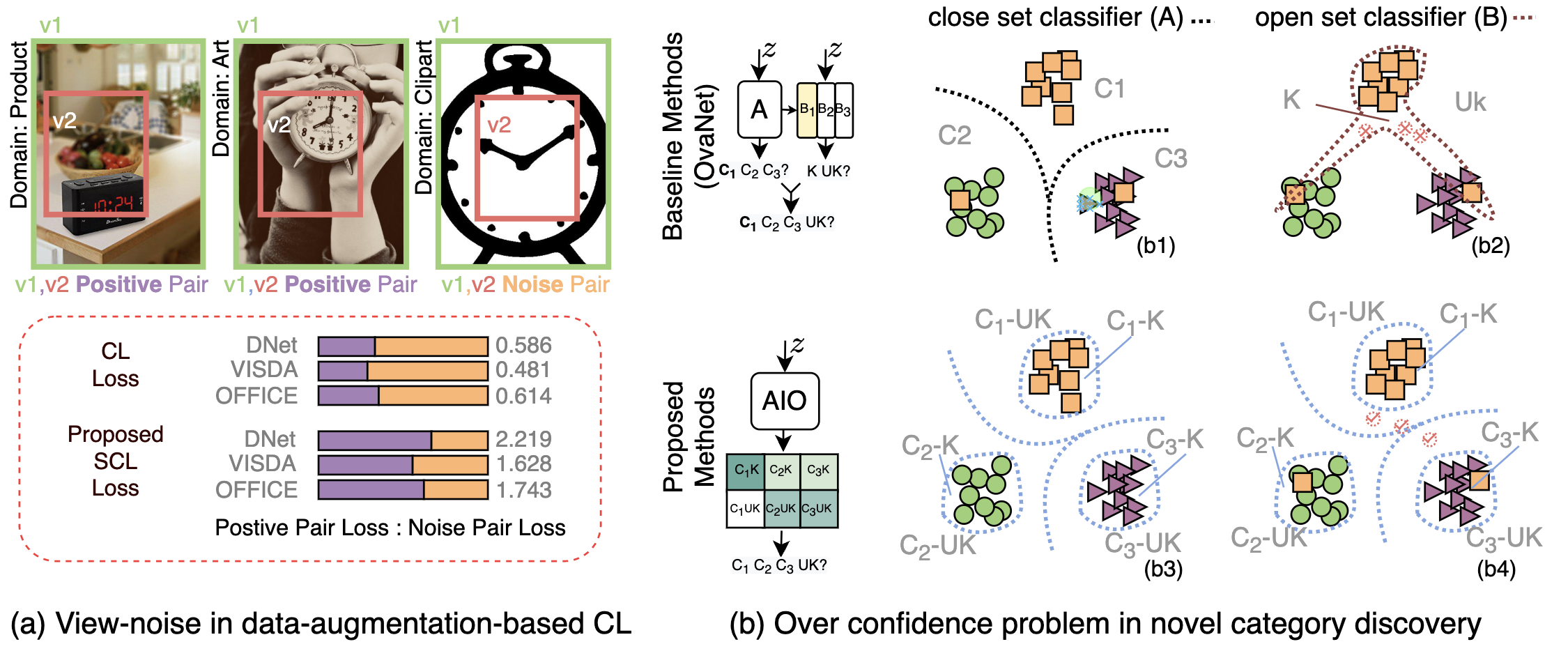}
  \vspace{-3mm}
\caption{
  \textbf{Problem demonstration and solutions.} 
  (a) View-noise problem in the backbone network fine-tuned by the CL. (a)-top shows the views generated by the same data augmentation scheme across three different domains. The difference in content style of the Clipart domain causes the regular data augmentation to produce views with vastly different semantics, producing noisy pairs.
  (b) Overconfidence problem of novel category classifiers. The dashed circle with a tick/cross means the test samples are classified correctly/incorrectly. }
  \label{fig_intro}
\vspace{-3mm}
\end{figure*}

The two main objectives of ODA and UNDA tasks are feature transfer and novel category discovery. However, the current methods have limitations in achieving both objectives, which hinders further improvement of these tasks. Specific problems are depicted in Fig.~\ref{fig_intro}.

\textbf{The view-noise problem in data augmentation affects the feature transfer.} 
Recently, data-augmentation-based contrastive learning~(CL) has been used for unsupervised fine-tune and has yielded excellent results in various downstream tasks~\cite{TingChen2020ASF, zbontar2021barlow, ZelinZang2022DLMEDL}. Therefore, several studies~\cite{QingYu2022SelfLabelingFF, LiangChen2022EvidentialNC} have attempted to enhance UNDA by introducing CL. However, these approaches are not based on data-augmentation-based CL schemes but are based on neighbor relationships of the original data.
Although view noise in CL has started to receive attention in network fine-tuning ~\cite{chuang2022robust}, the view noise problem in UNDA cannot be seen as similar to it.
Domain differences of data in UNDA introduce drastic and persistent view noise and cause more severe damage. 

As shown in the top of Fig.~\ref{fig_intro} (a), a more severe view-noise problem occurs if the same augmentation scheme is used in different domain data.
In detail, view~1 (v1) and view~2 (v2) are specific augmentation results across all doamins. In `product' domain and `art' domain, v1 and v2 have similar semantics, noted as positive pairs, but they have different semantics in the `clipart' domain, noted as noise pairs. The noise pairs contradict the accurate semantic information and, therefore, generate false gradients that corrupt the network training. More importantly, other data augmentation strategies, such as color perturbation, also lead to divergent semantic changes in different domains, leading to the view-noise problem. Addressing view-noise problem of cross-domain data augmentation training can further release the potential of CL on transfer learning.

\textbf{Overconfidence problem of classifiers~(closed-set classifier and open-set classifier) affects novel category recognition performance.}
Recently, OVANet~\cite{KuniakiSaito2021OVANetON} and its variants~\cite{YifanWang2022ExploitingIA} have received much attention. These methods combine closed-set classifiers and open-set classifiers to identify known and unknown classes. 
However, each of the open-set classifiers only corresponds to a single known class. 
When determining whether a sample belongs to a novel class, the open-set classifier does not compete with other open-set classifiers. It only determines whether the output of this classifier is greater than a certain threshold. This counter-intuitive strategy causes insufficient inter-class competition, which in turn leads to classifiers that are more likely to fall into overfitting and overconfidence. When label noise is present in the source domain, such data is almost inevitable and the harmful effects of overconfidence are amplified.
As shown at the top of Fig.~\ref{fig_intro} (b), even though the closed-set classifier is not affected by label-noise, the classification boundary of the open-set classifier can become very sharp due to label-noise and overfitting issues, which eventually causes the target domain samples to be misclassified.

To address the challenges mentioned above, we propose the Soft-contrastive All-in-one Network (denoted as \ours) for both UNDA and ODA.

\textbf{For view-noise problem}, we introduce a soft contrastive learning (SCL) loss. Unlike the commonly used contrastive learning (CL) loss, our SCL loss considers the similarity of views in the latent space to assess the reliability of the view. This enables us to construct a more effective loss function by incorporating reliability. In Fig.~\ref{fig_intro}(a), we compare our SCL loss to the CL loss in dealing with noise pair data and demonstrate that our SCL loss effectively reduces the influence of noise pairs on the model.

\textbf{For overconfidence problem of independent classifiers.} An all-in-one~(AIO) classifier is designed to replace the closed-set classifier and open-set classifier. 
The decision-making process of the AIO classifier is closer to that of humans. The AIO classifier assumes that identifying a sample belonging to a novel category requires determining that it does not belong to any known classes. Based on this assumption, a new loss function has been defined to train the AIO classifier. As shown in (b3) and (b4) of Figure~\ref{fig_intro}, as a result, the AIO classifier has smoother classification boundaries and reduces the adverse effects of label noise by introducing more comprehensive competition.

In experiments, we extensively evaluate our method on ODA and UNDA benchmarks and vary the proportion of unknown classes. The results show that the proposed \ourss outperforms all baseline methods on various datasets of the ODA and UNDA tasks.

\section{Related work}

\label{sec:related}
\textbf{Unsupervised Domain Adaptation~(UDA).}
The UDA~\cite{saenko2010} aims to learn a classifier for a target domain using labeled source data and unlabeled target data. UDA includes closed-set domain adaptation~(CDA), open-set domain adaptation~(ODA), partial domain adaptation~(PDA), and universal domain adaptation~(UNDA).
For CDA, we have $L_s = L_t$, where $L_s$ and $L_t$ are the label spaces of the source and target domains~\cite{ganin2014unsupervised, tzeng2017adversarial,long2015learning}.
For ODA~\cite{panareda2017open, saito2018open}, we have $|L_t - L_s| > 0$, $|L_t \cap L_s| = |L_s|$, and the presence of target-private classes in $|L_t - L_s|$.
For PDA, we have $|L_s - L_t| > 0$, $|L_t \cap L_s| = |L_t|$, and the presence of source-private classes in $|L_s - L_t|$.

\textbf{Universal Domain Adaptation~(UNDA).}
UNDA, also known as open-partial domain adaptation (OPDA) in some previous works, is proposed to handle the mixture of settings where $|L_s - L_t| > 0$ and $|L_t - L_s| > 0$ \cite{saito2020universal}. They emphasize the importance of measuring the robustness of a model to various category shifts since the details of these shifts cannot be known in advance.
In~\cite{UDA_2019_CVPR} and~\cite{saito2020universal}, a confidence score for known classes is computed, and samples with a score lower than a threshold are considered unknown.
The paper~\cite{bucci2020effectiveness} uses the mean of the confidence score as the threshold, implicitly rejecting about half of the target data as unknown. However, paper~\cite{saito2020universal} sets a threshold based on the number of classes in the source, which does not always work well. In a recent study, paper~\cite{YifanWang2022ExploitingIA} showed that exploiting inter-sample affinity can significantly improve the performance of UNDA. They propose a knowability-aware UNDA framework based on this idea.

\textbf{Contrastive learning based UNDA.}
Recently, contrastive learning~(CL), a kind of self-supervised learning paradigm~\cite{Xiao2020SelfsupervisedLG}, has achieved impressively superior performance in many computer vision tasks~\cite{TingChen2020ASF}. It aims to achieve instance-level discrimination and invariance by pushing semantically distinct samples away while pulling semantically consistent samples closer in the feature space~\cite{XinleiChen2020ExploringSS, XudongWang2021UnsupervisedFL}. 
Paper~\cite{LiangChen2022EvidentialNC} proposes to utilize mutual nearest neighbors as positive pairs to achieve feature alignment between the two domains.
Paper~\cite{LiangChen2022GeometricAC} constructs the random walk-based MNN pairs as positive anchors intra- and inter-domains and then proposes a cross-domain subgraph-level CL objective to aggregate local similar samples and separate different samples. To the best of our knowledge, no data-augmentation-based CL schemes are used to solve the UNDA problem.

\section{Methods}
\textbf{Notation.} In ODA and UNDA, we are given a source domain dataset $\mathcal{D}_{s}=\left\{\left(\mathbf{x}_{i}^{s}, {\widehat{y}_{i}}^{s}\right)\right\}_{i=1}^{N_{s}}$ and a target domain dataset $\mathcal{D}_{t} = \left\{\left(\mathbf{x}_{i}^{t} \right)\right\}_{i=1}^{N_{t}}$ which contains known categories and `unknown' categories. $L_s$ and $L_t$ denote the label spaces of the source and target respectively.
We assume that there is unavoidable noise and errors in the labels, so $\widehat{y}_{i}^{s}$ is noted as sampling from the real label ${y}_{i}^{s}$. The class-conditional random noise model is given by $P(\widehat{y}_{i}^{s} \neq {y}_{i}^{s}) = \rho^{s}$. We aim to label the target samples with either one of the $L_s$ labels or the `unknown' label. We train the model on $\mathcal{D}_{s} \cup \mathcal{D}_{t}$ and evaluate on $\mathcal{D}_{t}$.

\begin{figure*}
  \centering
  \includegraphics[width=0.999\linewidth]{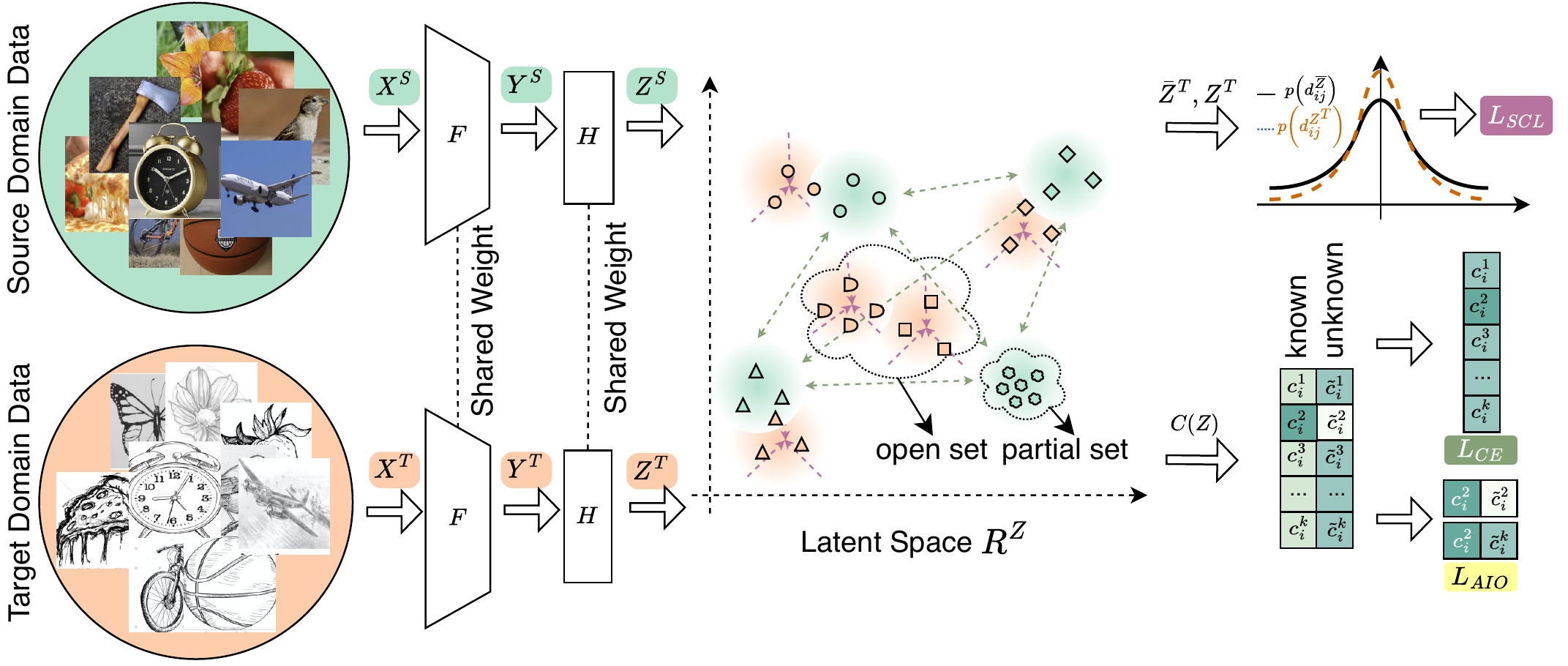}
  \caption{\textbf{Framework of \ours.} The proposed method includes a backbone network $F(\cdot)$, a projection head network $H(\cdot)$, and an all-in-one~(AIO) classifier $C(\cdot)$. The backbone network $F(\cdot)$ and projection head network $H(\cdot)$ map the source domain data $\mathbf{x}^{s}$ and the target domain data $\mathbf{x}^{t}$ into latent space.}
  \label{fig_main_training}
  \vspace{-6mm}
\end{figure*}

\textbf{Framework.} Fig.~\ref{fig_main_training} introduces the conceptual overview of \ours. The proposed method includes a backbone network $F(\cdot)$, a projection head network $H(\cdot)$, and an all-in-one~(AIO) classifier $C^\text{AIO}(\cdot)$. The backbone network $F(\cdot)$ and projection head network $H(\cdot)$ map the source domain data $\mathbf{x}^{s}_{i}$ and the target domain data $\mathbf{x}^{t}_{i}$ into latent space, $\mathbf{z}^{s}_{i}=H(\hat{z}^s_{i})=H(F(x^s_{i})), \ \ \ \mathbf{z}^{t}_{i}=H(\hat{z}^t_{i})=H(F(x^t_{i}))$.


\subsection{View-noise and Soft Contrastive Learning Loss}
Data-augmentation-based contrastive learning~(CL) involves binary classification over pairs of samples. Positive pairs are from the joint distribution $(x_i, x_j) \sim P_{x_i,x_j}$, labeled as $\mathcal{H}_{ij} = 1$, while negative pairs are from the product of marginals $(x_i, x_j) \sim P_{x_i}P_{x_j}$, labeled as $\mathcal{H}_{ij} = 0$. The CL learns representations by maximizing the similarity between positive samples and minimizing the similarity between negative samples using the InfoNCE loss~\cite{AaronvandenOord2018RepresentationLW}.
\begin{equation}
  \begin{aligned}
     & \mathcal{L}_\text{CL}\left(x_i, x_j, \{x_k \}_{k=1}^{N_K}\right)
    \\
      \!= &\!-\!\log\!\frac{\exp({z_i^T\!z_j})}{\sum_{k^=1}^{N_K}\!\exp({z_i^T\!z_k})}
    \!=\!-\!\log\!\frac{\exp(S(z_i,\!z_j))}{\sum_{k=1}^{N_K}\exp(S(z_i,\!z_k))}\!
    \label{eq_CL}
  \end{aligned}
\end{equation}
where $(x_i, x_j)$ is positive pair and $(x_i, x_k)$ is negative pair, and $z_i, z_j, z_k$ are the embedding of $x_i, x_j, x_k$, $N_K$ is the number of the negative pair. The similarity $S(z_i,z_j)$ is typically defined by cosine similarity.

The typical contrastive learning~(CL) loss assumes there is one positive sample and multiple negative samples. To design a smoother version of the CL loss, we transform it into a form based on positive and negative sample labels $\mathcal{H}_{ij}$. More details about the transformation from Eq.~(\ref{eq_CL}) to Eq.~(\ref{eq_nce2}) can be found in Appendix~\ref{app_proof_1}.
\begin{equation}
      \mathcal{L}_\text{CL}(\!x_i,\!\{\!x_k\!\}_{k=1}^{N_K}\!)
    \!=\!-\!\sum_{j=1}\{\! 
          \mathcal{H}_{ij}\!\log\!Q_{ij}\!+\!(\!1\!-\!\mathcal{H}_{ij}\!) \log\!\dot{Q}_{ij}
    \!\}\!
  \label{eq_nce2}
\end{equation}
where $\mathcal{H}_{ij}$ indicates whether $i$ and $j$ have been augmented from the same data. If $\mathcal{H}_{ij} = 1$, it means that $(x_i, x_j)$ is a positive pair, and if $\mathcal{H}_{ij} = 0$, it means that $(x_i, x_j)$ is a negative pair. The variable $Q_{ij}=\exp (S(z_i, z_j))$ represents the density ratio, which is defined in~\cite{AaronvandenOord2018RepresentationLW} and estimated by the backbone network.
In general, positive pairs are obtained through stochastic data augmentation, which means that the learning process inevitably introduces view-noise (shown in Fig.~\ref{fig_intro}(a)). As a result, view-noise introduces the wrong gradient, which can corrupt the network's training. Furthermore, finding suitable data augmentation schemes for all domains is challenging for UNDA data that exhibits vast domain variance.
This view-noise problem limits employing data-augmentation-based CL methods in UNDA. 

\textit{To address the view-noise problem described above}, we propose \textbf{Soft Contrastive Learning (SCL)}. SCL attenuates the negative impact of incorrect samples by assigning different weights to different positive and negative samples, which are estimated by calculating similarity through its own backbone network. The loss function of SCL is as follows:
\begin{equation}
  \begin{aligned}
      & \mathcal{L}_\text{scl}(x_i,\{x_j\}_{j=1}^{N_K})                                                          \\
    = & \!-\! {\sum_{j=1} \{  \! P_{ij} \log\left(Q_{ij}\right)\!+ (1\!-\!P_{ij}) \log\left(1\!-\!Q_{ij}\right)} \}, \\
  \end{aligned}
\end{equation}
where $P_{ij}$ is the weights, regarded as a soft version of $\mathcal{H}_{ij}$, and $Q_{ij}$ is density ratio.
\begin{equation}
  \begin{aligned}
    P_{ij}
           & = \left\{
    \begin{array}{lr}
      e^\alpha \kappa(y_i, y_j) \;\;\; \text{if} \;\; \mathcal{H}_{ij}=1 \\
      \kappa(y_i, y_j) \;\;\;\;\;\;\;  \text{otherwise}                       \\
    \end{array}
    \right.,                    \\
    Q_{ij} & = \kappa(z_i, z_j).
  \end{aligned}
\end{equation}
where hyper-parameter $\alpha \in [0,1]$ introduces prior knowledge of data augmentation relationship $\mathcal{H}_{ij}$ into the model training. To map the high-dimensional embedding vector~(such as $(y_i, y_j)$) to a probability value, a kernel function $\kappa(\cdot)$ is used. Commonly used kernel functions, including Gaussian kernel functions~\cite{keerthi2003asymptotic}, radial basis kernel functions~\cite{song2008research}, and t-distribution kernel functions~\cite{ZelinZang2022DLMEDL, zang2023udrn}, can be employed. In this paper, we use the t-distribution kernel function $\kappa^\nu(\cdot)$ because it exposes the degrees of freedom and allows us to adjust the closeness of the distribution in the dimensionality reduction mapping~\cite{BolianLi2021TrustworthyLC,Zang_tvcg_EVNET}. The t-distribution kernel function is defined as follows,
\begin{equation}
  \begin{aligned}
     & \kappa^\nu(z_i, z_j) =
    \frac
    {\Gamma\left(\frac{\nu +1}{2}\right)}
    {\sqrt{\nu  \pi} \Gamma\left(\frac{\nu }{2}\right)}
    \left(
    1+\frac{\|z_i- z_j\|_{2}^{2}}{\nu}
    \right)^{-\frac{\nu +1}{2}},
  \end{aligned}
  \label{eq_tkernal}
\end{equation}
where $\Gamma(\cdot)$ is the Gamma function. The degrees of freedom $\nu$ control the shape of the kernel function. The different degrees of freedom~($\nu^y, \nu^z$) is used in $\mathcal{R}^y$ and $\mathcal{R}^z$ for the dimensional reduction mapping.

SCL loss uses a softened optimization target, as opposed to the hard target of a typical CL loss, while avoiding the strong misresponse to noise labels. A formal discussion of the differences between the two losses is provided in Appendix~\ref{app_proof_3}. Furthermore, we can prove that SCL loss can maintain a higher signal-to-noise ratio when dealing with view-noise. See Appendix~\ref{app_proof_3} for more details.

\subsection{Overconfidence and All in One~(AIO) Classifier}

The current advanced UNDA methods~\cite{KuniakiSaito2021OVANetON, YifanWang2022ExploitingIA} combines a closed-set classifier $C^A$ and open-set classifiers $\{C^B_k\}_{k\in \mathcal{K}}$ to identify samples belonging to an unknown or specific known class, where $\mathcal{K}$ is the set of known classes $\mathcal{K} \in\{1, \ldots, N_K\}$ and $N_K$ is the number of known classes. The inference process consists of two steps. First, $C^A$ identifies the most likely target class($k$-th class). Second, the corresponding sub-classifiers $C^B_k$ determines whether the sample is a known or unknown class~(see Fig.~\ref{fig_intro} (b) baseline method).
In training a single open-set classifier $C^B_k$, samples with $y_i=k$ are defined as positive samples, while samples with $y_i \neq k$ are defined as negative samples. As a result, the open-set classifier can become overconfident by focusing on labels that only contain information from single classes, and ignoring the competing relationships of different known classes~\cite{YifanWang2022ExploitingIA}. This overconfidence is manifested in sharp category boundaries, and in failures to generalize from the source domain to the target domain. In addition, the noise in the labels compounds the damaging effects of the overconfidence problem.

\begin{table*}[ht]
  \addtolength{\tabcolsep}{-3pt}
  \caption{\textbf{H-score comparison of Office and DomainNet datasets in the UNDA setting.} Single \ourss indicates that uniform settings, and \ours$^\divideontimes$ indicates selecting the best hyperpatameters using the grid search. \textbf{Bolded} means best performance, \underline{underlined} means 2\% better than other methods. The brackets after the dataset indicate ($|L_s-L_t|$, $|L_t-L_s|$, $|L_s \cap L_t|$).}
  \vspace{-6mm}
  \begin{center}
    \begin{tabular}{l|l|cccccc|c|cccccc|c}
      \toprule
      \multirow{2}{*}{Method} & \multirow{2}{*}{REF}               & \multicolumn{6}{c|}{Office (10 / 10 / 11)} & \multirow{2}{*}{Avg}  & \multicolumn{6}{c|}{DomainNet (150 / 50 / 145))} & \multirow{2}{*}{Avg}                                                                                                                                                        \\
                              &                                    & A2D                                        & A2W                   & D2A                                              & D2W                   & W2D       & W2A                   &                       & P2R    & R2P       & P2S       & S2P       & R2S                   & S2R    &           \\\hline
      DANCE                   & NeurIPS2020  \cite{saito2020universal}                         & 78.6                                       & 71.5                  & 79.9                                             & 91.4                  & 87.9      & 72.2                  & 80.3                  & 21.0   & 47.3      & 37.0      & 27.7      & 46.7                  & 21.0   & 33.5      \\
      DCC                     & CVPR2021  \cite{GuangruiLi2021DomainCC}                         & 88.5                                       & 78.5                  & 70.2                                             & 79.3                  & 88.6      & 75.9                  & 80.2                  & 56.9   & 50.3      & 43.7      & 44.9      & 43.3                  & 56.2   & 49.2      \\

      OVANet                  & ICCV2021 \cite{KuniakiSaito2021OVANetON}                        & 85.8                                       & 79.4                  & 80.1                                             & 95.4                  & 94.3      & 84.0                  & 86.5                  & 56.0   & 51.7      & 47.1      & 47.4      & 44.9                  & 57.2   & 50.7      \\
      TNT                     & AAAI2022  \cite{LiangChen2022EvidentialNC}                         & 85.7                                       & 80.4                  & 83.8                                             & 92.0                  & 91.2      & 79.1                  & 85.4                  & ---    & ---       & ---       & ---       & ---                   & ---    & ---       \\
      GATE                    & CVPR2022   \cite{LiangChen2022GeometricAC}                        & 87.7                                       & 81.6                  & 84.2                                             & 94.8                  & 94.1      & 83.4                  & 87.6                  & ---    & ---       & ---       & ---       & ---                   & ---    & ---       \\
      D+SPA               & NeurIPS2022\cite{kundu2022subsidiary} & \textbf{90.4}                                       & 83.8                  & 83.1                                            & 90.5                  & 88.6      & 86.5                  & 87.2                  & \textbf{59.1}   & 52.7      & 47.6      & 45.4      & 46.9                  & 56.7   & 51.4      \\
\hline
      \ours                   & ours                               & 89.9                                  & \underline{\bf{87.6}} & \underline{\bf{87.6}}                            & \underline{\bf{98.4}} & \underline{\bf{98.3}} & {89.0}                & \underline{\bf{91.8}} & {57.4} & \bf{52.9} & \bf{47.9} & \bf{48.2} & \bf{47.0}             & {57.9} & \bf{52.0} \\
      \ours$^\divideontimes$  & ours                               & \bf{90.4}                      & \underline{\bf{89.9}} & \underline{\bf{87.8}}                            & \underline{\bf{98.9}} &  \underline{\bf{98.3}} & \underline{\bf{95.6}} & \underline{\bf{93.5}} & {57.8} & \bf{52.9} & \bf{47.9} &  \bf{48.4} & \bf{47.2} & {57.9} & \bf{52.1} \\
      \toprule
    \end{tabular}
  \end{center}
  \vspace{-6mm}
  \label{tb_office_domainnet_opda}
\end{table*}

\textit{To address the problem described above}, we attribute the cause to the inadequate competition of a single open-set classifier. Specifically, each open-set classifier only completes binary classification and neglects to observe more diverse labels. As a result, the classifier overfits and produces exceptionally sharp classification boundaries, guided by the simple learning task. Another important reason is that it is inconsistent with human common sense for open-set classifiers to consider only one known class when identifying new classes. Humans need to judge whether new classes belong to known classes before identifying them as new classes. To this end, we propose the All-in-One~(AIO) classifier $C^\text{AIO}\left( \cdot \right)$. The AIO classifier assigns two output neurons to each known category, representing if samples belong to the specific category or not respectively. The forward propagation of $C^\text{AIO}\left( \cdot \right)$ is,
\begin{equation}
  \mathcal{C}_{x_i} = \left\{c_{x_i}^k,\tilde{c}_{x_i}^k| k\in \mathcal{K} \right\} = \sigma\left( C^\text{AIO}\left(z_{x_i}\right)\right),
\end{equation}
The $c_{x_i}^k$ and $\tilde{c}_{x_i}^k$ are the probability of $x_i$ being identified as $k$-th category or not. The $\sigma(\cdot)$ is a `top\_n softmax' function to ensure $ \sum_{k\in \mathcal{T}^N} \{c_{x_i}^k+\tilde{c}_{x_i}^k\}=1$, $\mathcal{T}^N$ is the top $N=20$ item of $\mathcal{C}_{x_i}$. The `top\_n softmax' is employed to balance the loss scale of different category numbers~(check Appendix~\ref{app_Top_nsoftmax} for more details).

We propose two principles for designing an intuitive UNDA classifier to train the AIO classifier to solve the dilemma in the previous section.
\textbf{(a) If the classifier assigns sample $x_i$ to a known class $y^{s}$}, it needs to make sure that the sample does not belong to other known classes $c_{x_i}^{y^{s}} > \max\{c_{x_i}^k\}_{k\in \mathcal{K} /y^{s}}$, and does not belong to an unknown class, $c_{x_i}^{y^{s}} > \max\{\tilde{c}_{x_i}^k\}_{k\in \mathcal{K}}$.
\textbf{(b) If the classifier assigns sample $x_i$ to an unknown class,} it needs to confirm that the sample does not belong to all known classes, $\max \{\tilde{c}_{x_i}^k\}_{k\in \mathcal{K}} > \max \{{c}_{x_i}^k\}_{k\in \mathcal{K}}$.

Next, we combine the two principles to obtain the following objective. For a sample of the source domain,
\begin{equation}
  c_{x_i}^{y^{s}} > \max \{\tilde{c}_{x_i}^k\}_{k\in \mathcal{K}} > \max\{c_{x_i}^k\}_{k\in \mathcal{K} /y^{s}},
  \label{eq_principles}
\end{equation}
Based on Eq.~(\ref{eq_principles}), we formulate the loss function as,
\begin{equation}
  \begin{aligned}
    \mathcal{L}_\text{AIO}(x^{s},y^{s})
    =  - [ & log(c_{x_i}^{y^{s}}) + \min \{log(\tilde{c}_{x}^k)\}_{k\in \mathcal{K}/y^{s}}         \\
           & + log\left(c_{x_i}^{y^{s}} - \max \{\tilde{c}_{x_i}^k\}_{k\in \mathcal{K}}  \right) ],
  \end{aligned}
\end{equation}
where the first and second terms of $L_\text{AIO}$ maximize $c_{x_i}^{y^{s}}$ and $\{\tilde{c}_{x}^k\}_{k\in \mathcal{K}/ y^{s}}$, thus guarantee that they have sufficiently positive predictions and are larger than $\{c_{x_i}^k\}_{k\in \mathcal{K} /y^{s}}$. Also, the third term guarantees that $c_{x_i}^{y^{s}} > \max \{\tilde{c}_{x_i}^k\}_{k\in \mathcal{K}}$. Implicitly, $\{c_{x_i}^k\}_{k\in \mathcal{K} /y^{s}}$ is guided to have the lowest activation.

\subsection{Learning \& Inference}
\textbf{Learning.} We combine the SCL loss and AIO loss to train the SAN. The overall loss is
\begin{eqnarray}
  \begin{aligned}
    E_{(\mathbf{x}_{i}^{s}, y_{i}^{s})} 
    \{ 
     \mathcal{L}_\text{ce}(\mathbf{x}_{i}^{s}, y_{i}^{s}) \!+\!
      \beta \mathcal{L}_\text{AIO} (\mathbf{x}_{i}^{s}, y_{i}^{s}) 
    \}+
      E_{\mathbf{x}_{i}^{t}} 
    \{ 
      \lambda \mathcal{L}_\text{scl}(\mathbf{x}_{i}^{t})
    \}.
    \label{eq:main}
  \end{aligned}
\end{eqnarray}
where $\mathcal{L}_\text{ce}$ represents the cross entropy loss. The network parameters are optimized by minimizing this loss. The hyperparameters $\lambda$ and $\beta$ are weighted. 
In comparison to existing ODA and UNDA methods~\cite{saito2020universal,kundu2022subsidiary}, our proposed method does not require more hyperparameters or loss functions. Instead, we design new feature alignment and classifier training schemes based on our theoretical analysis.

\textbf{Inference.}
Based on the fine-tuned model, the recognition results of the target domain can be obtained by forward propagation of the network. If $c_{x_i}^k$ is greater than others, then the sample $x_i$ is identified as a known class $k$. If any of $\{\tilde{c}_{x_i}^k\}_{k\in \mathcal{K}}$ of the AIO classifier achieves the maximum value, then the sample is identified as an unknown class.

\section{Results}
We evaluate our method in ODA and UNDA settings along with ablation studies. In addition, we assess the robustness with respect to the change of unknown target category size by varying the number of unknown categories.

\begin{table*}[t]
  \caption{\textbf{H-score comparison of OfficeHome datasets in the UNDA setting. (D+SPA means DCC+SPA)} Single \ours \space indicates that uniform settings, and \ours$^\divideontimes$ indicates selecting the best hyperparameters using the grid search. \textbf{Bolded} means best performance, \underline{underlined} means 2\% better than others.}
  \vspace{-6mm}
  \begin{center}
    \begin{tabular}{l|l|p{15pt}p{15pt}p{15pt}p{15pt}p{15pt}p{15pt}p{15pt}p{15pt}p{15pt}p{15pt}p{15pt}p{15pt}|c}
      \toprule
      \multirow{2}{*}{Method} & \multirow{2}{*}{REF}                       & \multicolumn{12}{c|}{OfficeHome (10 / 5 / 50)} &                                                                                                                                                                                                                               \\
                              &                                            & A2C                                            & A2P                   & A2R                   & C2A       & C2P                   & C2R           & P2A                   & P2C       & P2R                   & R2A       & R2C           & R2P       & Avg                   \\ \hline
      DANCE                   & NeurIPS2020  \cite{saito2020universal}     & 61.0                                           & 60.4                  & 64.9                  & 65.7      & 58.8                  & 61.8          & 73.1                  & 61.2      & 66.6                  & 67.7      & 62.4          & 63.7      & 63.9                  \\
      DCC                     & CVPR2021  \cite{GuangruiLi2021DomainCC}    & 58.0                                           & 54.1                  & 58.0                  & 74.6      & 70.6                  & 77.5          & 64.3                  & 73.6      & 74.9                  & \bf{81.0} & 75.1          & 80.4      & 70.2                  \\
      OVANet                  & ICCV2021 \cite{KuniakiSaito2021OVANetON}   & 62.8                                           & 75.6                  & 78.6                  & 70.7      & 68.8                  & 75.0          & 71.3                  & 58.6      & 80.5                  & 76.1      & 64.1          & 78.9      & 71.8                  \\
      TNT                     & AAAI2022  \cite{LiangChen2022EvidentialNC} & 61.9                                           & 74.6                  & 80.2                  & 73.5      & 71.4                  & 79.6          & 74.2                  & 69.5      & 82.7                  & 77.3      & 70.1          & 81.2      & 74.7                  \\
      GATE                    & CVPR2022   \cite{LiangChen2022GeometricAC} & 63.8                                           & 75.9                  & 81.4                  & 74.0      & 72.1                  & 79.8          & 74.7                  & 70.3      & 82.7                  & 79.1      & 71.5          & \bf{81.7} & 75.6                  \\

      D+SPA                   & NeurIPS2022 \cite{kundu2022subsidiary}     & 59.3                                           & 79.5                  & 81.5                  & \bf{74.7} & 71.7                  & \textbf{82.0} & 68.0                  & \bf{74.7} & 75.8                  & 74.5      & \textbf{75.8} & 81.3      & 74.9                  \\
      \hline
      \ours                   & ours                                       & \underline{\bf{66.7}}                          & 79.4                  & \underline{\bf{86.6}} & 73.2      & \underline{\bf{73.0}} & 79.5          & \underline{\bf{75.7}} & 64.0      & 82.6                  & 79.4      & 66.8          & 80.0      & \bf{75.9}             \\
      \ours$^\divideontimes$  & ours                                       & \underline{\bf{68.2}}                          & \underline{\bf{80.6}} & \underline{\bf{86.7}} & 73.4      & \underline{\bf{73.0}} & 79.8          & \underline{\bf{76.5}} & 64.9      & \underline{\bf{83.3}} & 80.1      & 67.1          & 80.1      & \underline{\bf{76.1}} \\
      \bottomrule
    \end{tabular}
  \end{center}
  \vspace{-4mm}
  \label{tb_officehome_opda}
\end{table*}

\textbf{Datasets.}
We utilize popular datasets in DA: Office~\cite{saenko2010}, OfficeHome~\cite{venkateswara2017Deep}, VisDA~\cite{peng2017visda}, and DomainNet~\cite{peng2018moment}. Unless otherwise noted, we follow existing protocols~\cite{KuniakiSaito2021OVANetON} to split the datasets into source-private ($|L_s-L_t|$), target-private ($|L_t-L_s|$) and shared categories ($|L_s \cap L_t|$).

\textbf{Baselines.}\quad
We aim to compare methods of universal domain adaptation~(UNDA), which can reject unknown samples, such as, CMU~\cite{fu2020learning}, DANCE~\cite{saito2020universal}, DCC~\cite{GuangruiLi2021DomainCC},  OVANet~\cite{KuniakiSaito2021OVANetON}, TNT~\cite{LiangChen2022EvidentialNC}, GATE~\cite{LiangChen2022GeometricAC} and D+SPA~\cite{kundu2022subsidiary}. 
We are looking at some contemporaneous work such as KUADA~\cite{YifanWang2022ExploitingIA}, UACP~\cite{YunyunWang2022TowardsAU} and UEPS~\cite{YifanWang2022ANF}, which we did not include in the comparison because the source code was not available and some of these works were not peer-reviewed. Instead of reproducing the results of these papers, we directly used the results reported in the papers with the same configuration. 

\textbf{Implementation.}\quad
Following previous works, such as OVANet~\cite{KuniakiSaito2021OVANetON} and GATE\cite{LiangChen2022GeometricAC}, we employ ResNet50~\cite{he2016deep} pre-trained on ImageNet~\cite{imagenet} as our backbone network. We train our models with inverse learning rate decay scheduling. The performance of the proposed \ourss in uniform settings is listed in the penultimate row of the table. A grid hyperparameter search is performed for each setup, and the optimal results obtained are marked with $^\divideontimes$. The selected hyperparameters for searching include $\lambda$, $\beta$, and $\alpha$.  For all experiments, $\nu^y=100$ and $\nu^z=10$. The network $H(\cdot)$ uses a two-layer MLP network with 2048 neurons. In summary, our method outperforms or is comparable to the baseline method in all different settings. More details of the implementation are in the Appendix.

\begin{table}[tbp]
    \caption{\textbf{H-score of Office datasets in the ODA setting.}}
    \vspace{-4mm}
    \begin{tabular}{l|p{14pt}p{14pt}p{14pt}p{14pt}p{14pt}p{18pt}|p{14pt}}
      \toprule
      \multirow{2}{*}{Method} & \multicolumn{6}{c|}{Office (10 / 0 / 11)} & \multirow{2}{*}{Avg}                                                             \\
                              & A2D                                       & A2W                  & D2A       & D2W       & W2D       & W2A       &           \\\hline
      DANCE                   & 84.9                                      & 78.8                 & 79.1      & 78.8      & 88.9      & 68.3      & 79.8      \\
      DCC                     & 58.3                                      & 54.8                 & 67.2      & 89.4      & 80.9      & 85.3      & 72.6      \\

      OVANet                  & \bf{90.5}                                 & 88.3                 & 86.7      & 98.2      & 98.4      & 88.3      & 91.7      \\
      TNT                     & 85.8         & 82.3         & 80.7         & 91.2         & 96.2         & 81.5         & 86.3 \\
      GATE                    & 88.4                                      & 86.5                 & 84.2      & 95.0      & 96.7      & 86.1      & 89.5      \\
      D+SPA                 & 92.3                                     & 91.7                 & 90.0      & 96.0      & 97.4      & 91.5      & 93.2      \\
   
      \hline
      \ours                   & \bf{90.5}                                 & \bf{93.5}            & {91.7}    & \bf{98.9} & \bf{100}  & \bf{92.8} & \bf{94.6} \\
      \ours$^\divideontimes$  & \bf{90.5}                                 & \bf{93.8}            & \bf{92.7} & \bf{99.3} & \bf{100}  & \bf{93.7} & \bf{95.0} \\
      \toprule
    \end{tabular}
    \vspace{-4mm}
    \label{tb_office_oda}
\end{table}

\begin{table}[tbp]
    \caption{\textbf{H-score of VisDA in UNDA and ODA setting.} }
    \vspace{-4mm}
    \begin{center}
    \begin{tabular}{l|c|c}
      \toprule
      \multirow{2}{*}{Method} & VisDA       & VisDA       \\
                              & ODA         & UNDA        \\
                              & (6 / 0 / 6) & (6 / 3 / 3) \\\hline
      CMU                     & 54.2        & 34.6        \\
      DANCE                   & 67.5        & 42.8        \\
      DCC                     & 59.6        & 43.0        \\
      OVANet                  & 66.1        & 53.1        \\
      TNT                     & 71.6        & 55.3         \\ 
      GATE                    & 70.8        & 56.4        \\
      
      \hline
      \ours                   & \bf{72.0}   & \bf{60.1}   \\
      \bottomrule
    \end{tabular}
    \end{center}
    \vspace{-10mm}
    \label{tb_VisDA_opda_oda}
\end{table}

\begin{figure*}[t]
  \centering
  \includegraphics[width=0.999\linewidth]{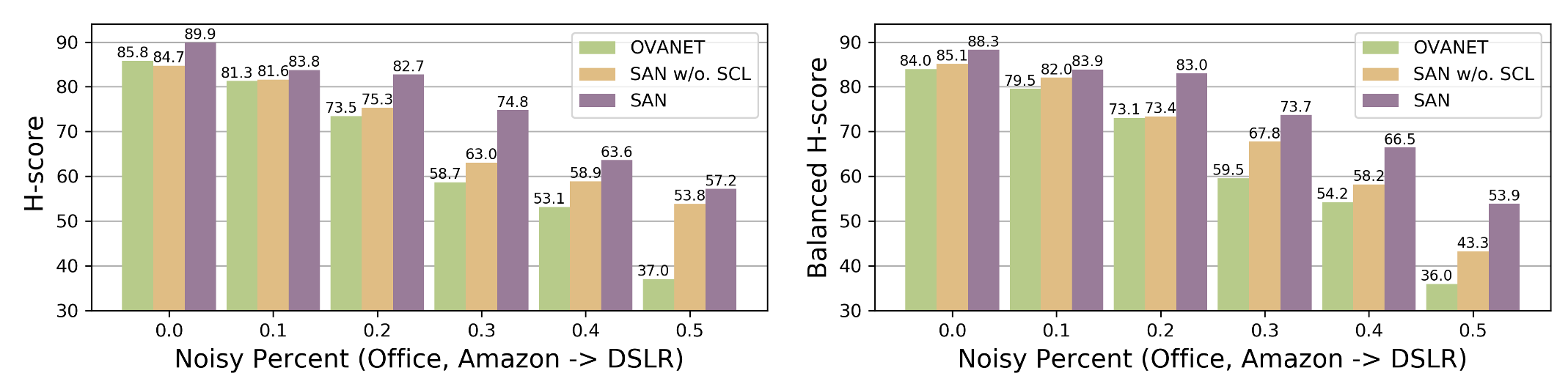}
  \vspace{-8mm}
  \caption{\textbf{Ablation study. OVANet v.s. \ourss w/o. SCL v.s. \ours.}
    H-score and Balance H-score Comparisons of Office datasets in the UNDA setting. The horizontal coordinate indicates the addition of a specified percentage of noise to the original domain, and the vertical coordinate indicates the performance of the different methods.
  }
  \label{fig_ablation_study}
  \vspace{-6mm}
\end{figure*}

\textbf{Evaluation Metric.}\quad
The H-score is usually used to evaluate standard or ODA methods because it considers the trade-off between the accuracy of known and unknown classes~\cite{bucci2020effectiveness}. 
H-score is the harmonic mean of the accuracy on common classes $A_c$ and the accuracy on “unknown” classes $A_t$, $ \text{H-score} = (2 A_c \cdot A_t)/(A_c+ A_t) $. The evaluation metric is high only when both the $A_c$ and $A_t$ are high. So, H-score can measure both accuracies of UNDA methods well. 
However, we find concerns about the fairness of the Hscore when the sample sizes of the known and unknown classes of the dataset differ significantly. 
For example, when the number of samples in the unknown category is much larger than the known category~(e.g., the Office-Home dataset), pairing one more sample from the known category leads $A_c$ to increase more significantly than the unknown category. Moreover, if $A_c$ increases, the H-score will greatly increase, which leads to unfairness about the known and unknown categories.
So, to achieve a higher h-score, the model will sacrifice the unknown category's accuracy to exchange for the common category's accuracy, which is unfair and impracticable in the real world.
Therefore, inspired by the idea of Weighted Harmonic Means~\cite{kanas2017weighted}, we propose the Balance H-score as a more equitable metric (the proof is shown in Appendix~\ref{app_proof_2}). For the dataset where the number of unknown categories in the sample is $ \theta $ times the number of common, we define $ \text{Balence H-score} =(1+\theta) A_c \cdot A_t/(\theta A_c+A_t) $.
This paper selects the Hscore as an evaluation metric for convenient comparison with the baseline approach. Meanwhile, the Balance H-score is used in the more profound analysis of the relative advantages of the proposed method.

\begin{figure*}[t]
  \centering
  \includegraphics[width=0.999\linewidth]{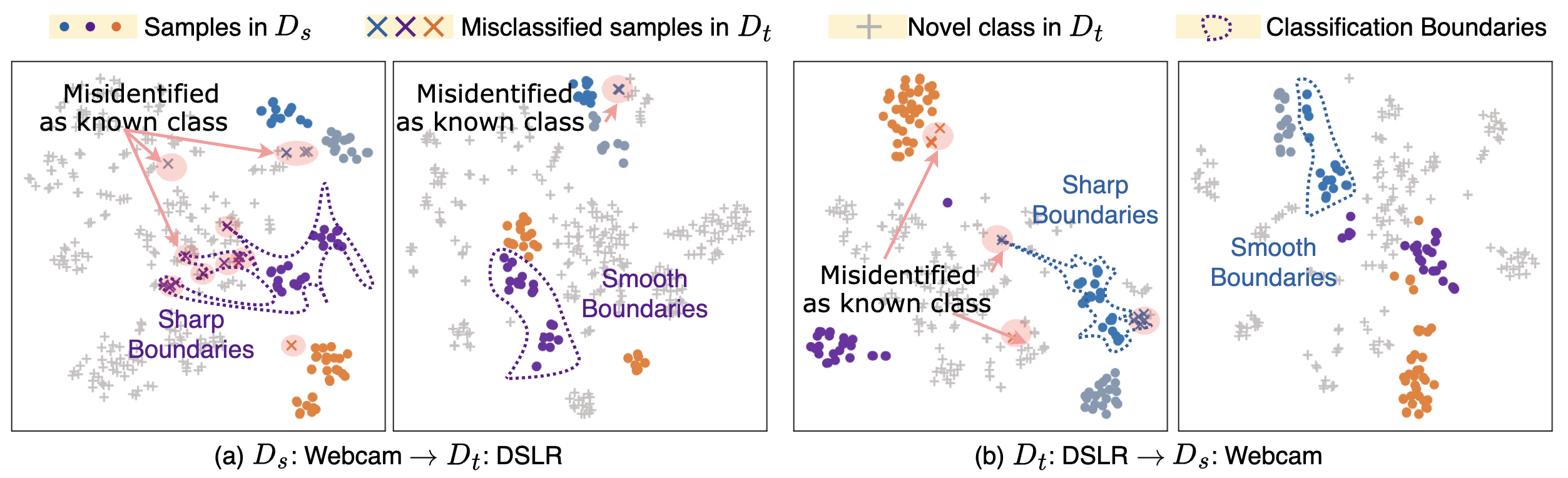}
  \vspace{-8mm}
  \caption{\textbf{Feature Visualization, OVANet V.S. \ours~(colors represent categories).}
  The t-SNE embeddings visualization of backbone network are shown in (a) and (b). For (a), source domain is webcam and the target domain is dslr. For (b), source domain is dslr and the target domain is webcam. We observed that the OVANet make more mistakes due to the overconfidence classification boundaries, while \ourss is better. We attribute this improvement to the fact that \ourss overcomes the problem of overconfidence.}
  \label{fig_exp2}
  \vspace{-5mm}
\end{figure*}

\textbf{Performance comparisons on UNDA setting.} \quad
From the results in Table~\ref{tb_office_domainnet_opda}, Table~\ref{tb_officehome_opda}, and Table~\ref{tb_VisDA_opda_oda}, \ourss achieves a new state-of-the-art~(SOTA) on all four datasets in the most challenging UNDA setting. Concerning H-score, \ourss outperforms the previous SOTA UNDA method on Office by 4.2\% and on OfficeHome by 0.3\%. On large-scale datasets, \ourss also gives more than 0.6\% improvement on DomainNet and more than 3.7\% on VisDA compared to all other methods in terms of H-score. In VisDA and DomainNet, the number of samples and/or classes differs greatly from those of Office and OfficeHome. 

\begin{figure*}[t]
  \includegraphics[width=0.999\linewidth]{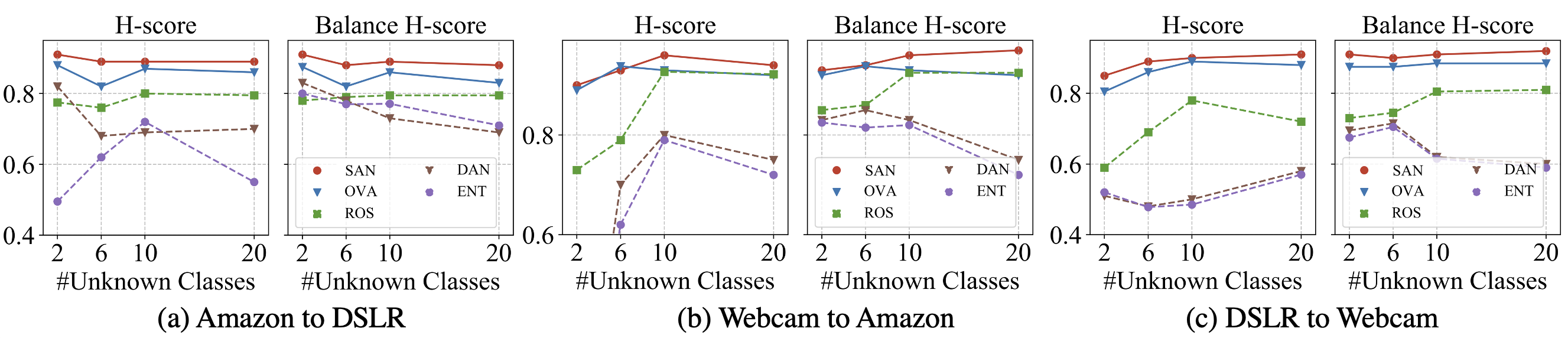}
  \vspace{-7mm}
  \caption{\textbf{H-score and Balance H-score comparison of Office dataset in ODA.}
    We vary the number of unknown classes using Office ($L_s \cap L_t| = 10$, $|L_s-L_t| = 0$). 
    The left and right parts, respectively, show H-score and Balance H-score. \ourss shows stable performance across different unknown class numbers, while baselines degrade performance in some settings.
  }
  \label{fig_exp3}
  \vspace{-4mm}
\end{figure*}

\textbf{Performance Comparisons on ODA setting.} \quad
For the ODA setting, the H-score comparison results are presented in Table~\ref{tb_office_oda} and Table~\ref{tb_VisDA_opda_oda}. Our method performs better than all the UNDA baselines on Office and VisDA datasets, with 1.4\% and 1.2\% H-score improvement. 

\textbf{Overview of Results.} \quad
Under these two scenarios with "unknown" samples, \ourss shows a more robust capability for separating common and private categories, which benefits from the SCL loss function and AIO classifier. Compared with GATE, a previous SOTA method tailored for the ODA setting, \ourss is also superior on all datasets. This evidence shows that \ourss gains a better trade-off between common categories classification and private samples identification. 

\begin{table}[t]
  \addtolength{\tabcolsep}{-3pt}
  \caption{\textbf{Ablation study.} H-score comparison on all four datasets in UNDA setting.}
  \vspace{-3mm}
  \centering
  \small
  \begin{tabular}{l|cccccc|c|cccccc|c}
    \toprule
    \multirow{2}{*}{Method} & \multirow{1}{*}{Office}     & \multirow{1}{*}{OfficeHome} & \multirow{1}{*}{DomainNet}    & \multirow{1}{*}{VisDA}   \\
                            & \multirow{1}{*}{(10/10/11)} & \multirow{1}{*}{(10/5/50)}  & \multirow{1}{*}{(150/50/145)} & \multirow{1}{*}{(6/3/3)} \\
    \midrule

    \textit{\ours}          & \bf{91.8}                   & \bf{75.9}                   & \bf{52.0}                     & \bf{60.1}                \\
    \textit{w/o. AIO}       & {90.5}                      & {74.6}                      & {51.0}                        & {57.2}                   \\
\textit{w/o. SCL}       & {78.9}                      & {73.0}                      & {50.7}                        & {55.2}                   \\
    \textit{w. CL}   & {78.2}                      & {73.3}                      & {50.5}                        & {52.7}                   \\
    \textit{OVANet}         & {77.9}                      & {71.8}                      & {50.7}                        & {53.1}                   \\
    \toprule
  \end{tabular}
  \vspace{-6mm}
  \label{tb_ablation}
\end{table}

\subsection{Analysis in Universal Domain Adaptation}
\textbf{Ablation study, the effect of SCL Loss.}
We conduct controlled experiments to verify the necessity of the soft contrastive learning~(SCL) Loss on all four datasets in UNDA settings, and the results are shown in Table~\ref{tb_ablation}. The \underline{\textit{\ourss}} shows the performance of the proposed method. The \underline{\textit{w/o. SCL}} shows the performance of the SCL loss with $\mathcal{L}_{SCL}(\mathbf{x}_{i}^{t}, {y_{i}}^{t})$ removed from the overall loss of \ours. The \underline{\textit{w. CL}} shows the performance of the SCL loss with $\mathcal{L}_{SCL}(\mathbf{x}_{i}^{t}, {y_{i}}^{t})$ replaced by the CL loss $\mathcal{L}_\text{CL}$ in Eq.~(\ref{eq_CL}). 
The \underline{\textit{OVANet}} shows the performance of OVANet.
The above experiences indicate that the SCL Loss significantly outperforms typical CL loss. We attribute the failure of $\mathcal{L}_\text{CL}$ to the fact that the view-noise caused by domain bias cannot be ignored. In addition, SCL loss can better alleviate this problem. 

\textbf{Ablation study, the effect of AIO classifier.}
We further conduct controlled experiments to verify the necessity of the All in one~(AIO) classifier. 
In Table.~\ref{tb_ablation}, the \underline{\textit{w/o. AIO}} shows the performance of the AIO classifier replaced by the open-set and closed-set classifier. The control experiments on all four datasets indicate that the AIO Classifier brings improvements. The improvement from the AIO classifier is not as significant as that from SCL, probably because the label noise in the dataset is not significant. We further verify this idea by manually adding some label noise, and the experiment results are shown in Fig.~\ref{fig_ablation_study}. The results show that the \underline{\textit{\ourss}} and \underline{\textit{\ourss w/o. SCL}} exceed the baseline more significantly as the proportion of noise increases. 

\textbf{The overconfidence problem and its mitigation by \ours.}
Many current approaches are based on a combination of open-set classifiers and closed-set classifiers. We consider that they fail to achieve further improvements because the strategy of open-set classifiers leads to overconfidence. One direct evidence is that \ourss achieves a more significant advantage in datasets with fewer samples (e.g., Office). To explore the adverse effects of overconfidence, we perform a visual analysis of the W2D and D2W settings of the Office dataset in Fig.~\ref{fig_exp2}. We find that the open set classifier of OVANet make mistake in some novel class in target domain~(e.g. the scatters marked by cross). It can be attributed to the overconfidence classifier causing the over-sharp classification boundaries thus wrong testing results are output. We consider this is the result of overconfidence. Contrastingly, the same class is handled well by \ours.

\textbf{The effect of the proportion of unknown samples on H-score, and the advantage of \ourss on Balance H-score.}
H-score introduces fairness bias if there is a large quantitative difference between the sample size of unknown and known. To explore the fairness of the H-score, we changed the number of unknown classes in the target domain and then tested the performance of the H-score and balance H-score~(in Fig.~\ref{fig_exp3}). The results show that changing the number of unknown classes dramatically changes the H-score. In contrast, the balance H-score exhibits higher stability. This suggests that the balance H-score is a more stable indicator for the proportion of unknown class samples. Its fairness is demonstrated in Appendix~\ref{app_proof_2}. In addition, Fig.\ref{fig_ablation_study} and Fig.\ref{fig_exp3} show that the proposed \ourss has more evident advantages in both the H-score and Balance H-score.

\section{Conclusion}
ODA and UNDA tasks aim to transfer the knowledge learned from a label-rich source domain to a label-scarce target domain without any constraints on the label space. In this paper, to solve the view-noise problem of data-augmentation-based CL and the overconfidence problem of novel category classifier, a framework named Soft-contrastive All-in-one Network~(\ours) is proposed. \ourss includes SCL loss which can avoid the over-response of typical CL loss and enable data augmentation-based contrastive loss to improve the performance of ODA and UNDA. In addition, \ourss includes an all-in-one~(AIO) classifier to improve the robustness of novel category discovery. Extensive experimental results on UNDA and ODA demonstrate the advantages of \ourss over existing methods.

\section*{Acknowledgement}
This work was supported by National Key R\&D Program of China~(No.2022ZD0115100), National Natural Science Foundation of China Project~(No. U21A20427), and Project~(No. WU2022A009) from the Center of Synthetic Biologyand Integrated Bioengineering of Westlake University. This work was supported by Alibaba Group through Alibaba Innovative Research Program. We thank the Westlake University HPC Center for providing computational resources.

{\small
   \bibliographystyle{ieee_fullname}
   \bibliography{egbib,bib_ova}
}

\clearpage
\appendix
\onecolumn

\

\begin{center}
{\Large
Supplementary Materials:\\
Boosting Novel Category Discovery Over Domains\\
with Soft Contrastive Learning and All-in-One Classifier \\}
\vspace{5mm}
{
}
\end{center}

\vspace{10mm}

\section{Details of SCL loss}
\label{app_proof_1}

\subsection{Details of the transformation from Eq.~(\ref{eq_CL}) to Eq.~(\ref{eq_nce2})}

We start with $L_\text{CL} =  - \log \frac{ \exp(S(z_i, z_j))}{\sum_{k=1}^{N_K} \exp(S(z_i, z_k))}$ (Eq.~(\ref{eq_CL})), then

$$ L_\text{CL} = \log N_K - \log \frac{ \exp(S(z_i, z_j))}{\frac{1}{N_K}\sum_{k=1}^{N_K} \exp(S(z_i, z_k))}. $$

We are only concerned with the second term that has the gradient. Let $(i,j)$ are positive pair and $(i,k_1), \cdots, (i,k_N) $ are negative pairs. The overall loss associated with point $i$ is:
\begin{equation*}
  \begin{aligned}
      & - \log \frac
    {\exp(S(z_i, z_j))}
    { \frac{1}{N_K} \sum_{k=1}^{N_K} \exp(S(z_i, z_k))}                                                          \\
    = & - \left[
      \log \exp(S(z_i, z_j)) - \log
    { \frac{1}{N_K} \sum_{k=1}^{N_K} \exp(S(z_i, z_k))} \right]                                                  \\
    = & - \left[
      \log \exp(S(z_i, z_j)) - \sum_{k=1}^{N_K} \log \exp(S(z_i, z_{k})) +
    \sum_{k=1}^{N_K} \log \exp(S(z_i, z_{k})) - \log { \frac{1}{N_K} \sum_{k=1}^{N_K} \exp(S(z_i, z_k))} \right] \\
    = & - \left[
      \log \exp(S(z_i, z_j)) - \sum_{k=1}^{N_K} \log \exp(S(z_i, z_{k})) +
    \log \Pi_{k=1}^{N_K}  \exp(S(z_i, z_{k})) - \log { \frac{1}{N_K} \sum_{k=1}^{N_K} \exp(S(z_i, z_k))} \right] \\
    = & - \left[
      \log \exp(S(z_i, z_j)) - \sum_{k=1}^{N_K} \log \exp(S(z_i, z_{k})) +
    \log \frac {\Pi_{k=1}^{N_K}  \exp(S(z_i, z_{k}))}{ \frac{1}{N_K} \sum_{k=1}^{N_K} \exp(S(z_i, z_k))} \right] \\
  \end{aligned}
\end{equation*}

We focus on the case where the similarity is normalized, $S(z_i, z_k) \in [0,1]$. The data $i$ and data $k$ is the negative samples, then $S(z_i, z_k)$ is near to $0$, $\exp(S(z_i, z_{k}))$ is near to $1$, thus the $\frac {\Pi_{k=1}^{N_K} \exp(S(z_i, z_{k}))}{ \frac{1}{N} \sum_{k=1}^{N_K} \exp(S(z_i, z_k))}$ is near to 1, and $\log \frac {\Pi_{k=1}^{N_K}  \exp(S(z_i, z_{k}))}{ \frac{1}{N} \sum_{k=1}^{N_K} \exp(S(z_i, z_k))}$ near to 0. We have

\begin{equation*}
  \begin{aligned}
    L_\text{CL}
     & \approx  - \left[
    \log \exp(S(z_i, z_j)) - \sum_{k=1}^{N_K} \log \exp(S(z_i, z_{k})) \right] \\
  \end{aligned}
\end{equation*}

We denote $ij$ and $ik$ by a uniform index and use $\mathcal{H}_{ij}$ to denote the homology relation of $ij$.

\begin{equation*}
  \begin{aligned}
    L_\text{CL}
     & \approx - \left[
    \log \exp(S(z_i, z_j)) - \sum_{k=1}^{N_K} \log \exp(S(z_i, z_{k})) \right]                                       \\
     & \approx - \left[
    \mathcal{H}_{ij} \log \exp(S(z_i, z_j)) - \sum_{j=1}^{N_K} (1-\mathcal{H}_{ij}) \log \exp(S(z_i, z_{j})) \right] \\
     & \approx - \left[
      \sum_{j=1}^{N_K+1} \left\{  \mathcal{H}_{ij} \log \exp(S(z_i, z_j)) +  (1-\mathcal{H}_{ij}) \log \{\exp(-S(z_i, z_{j}))\}
      \right\}
    \right]                                                                                                          \\
  \end{aligned}
\end{equation*}

we define the similarity of data $i$ and data $j$ as $Q_{ij} = \exp(S(z_i, z_j))$ and the dissimilarity of data $i$ and data $j$ as $\dot{Q}_{ij} =  \exp(-S(z_i, z_j))$.

\begin{equation*}
  \begin{aligned}
    L_\text{CL} \approx - \left[
      \sum_{j=1}^{N_K+1} \left\{  \mathcal{H}_{ij} \log Q_{ij} +  (1-\mathcal{H}_{ij}) \log \dot{Q}_{ij}
      \right\}
    \right] \\
  \end{aligned}
\end{equation*}


\subsection{The proposed SCL loss is a smoother CL loss}

This proof tries to indicate that the proposed SCL loss is a smoother CL loss. We discuss the differences by comparing the two losses to prove this point. 
the forward propagation of the network is,
${z}_{i}=H(\hat{z}_{i}), \hat{z}_{i} =F(x_{i})$, 
${z}_{j}=H(\hat{z}_{j}), \hat{z}_{j} =F(x_{j})$.
We found that we mix $y$ and $\hat{z}$ in the main text, and we will correct this in the new version. So, in this section 
${z}_{i}=H(y_{i}), y_{i} =F(x_{i})$, 
${z}_{j}=H(y_{j}), y_{j} =F(x_{j})$ is also correct.

Let $H(\cdot)$ satisfy $K$-Lipschitz continuity, then
$
  d^z_{ij} = k^* d^y_{ij} , k^* \in [1/K, K],
$
where $k^*$ is a Lipschitz constant. The difference between $L_\text{SCL}$ loss and $L_\text{CL}$ loss is,
\begin{equation}
  \begin{aligned}
    L_{\text{CL}}- L_\text{SCL} \approx
    \sum_j \biggl[
    \left(
    \mathcal{H}_{ij} - [1+(e^\alpha -1)\mathcal{H}_{ij}] \kappa \left(d^y_{ij} \right)
    \right)
    \log
    \left(
    \frac
    {1}
    {\kappa \left( d_{ij}^{z}\right)}
    -
    1
    \right)
    \biggl] .
  \end{aligned} \label{eq_sclcl}
\end{equation}
Because the  $\alpha > 0$, the proposed SCL loss is the soft version of the CL loss. if $\mathcal{H}_{ij}=1$, we have:

\begin{equation}
  \begin{aligned}
    (L_{\text{CL}} - L_{\text{SCL}})  |_{\mathcal{H}_{ij} =1} = \sum
    \biggl[
      \left(
      (1 - e^\alpha) \kappa \left( k^* d^z_{ij} \right)
      \right)
      \log
      \left(
      \frac
      {1}
      {\kappa \left( d_{ij}^{z}\right)}
      -
      1
      \right)
    \biggl] \\
  \end{aligned}
\end{equation}

then:

\begin{equation}
  \begin{aligned}
       \lim_{\alpha \to 0}
    ( L_{\text{CL}} - L_{\text{SCL}} ) |_{\mathcal{H}_{ij} =1}
    = \lim_{\alpha \to 0} \sum
    \biggl[
      \left(
      (1 - e^\alpha) \kappa \left( k^* d^z_{ij} \right)
      \right)
      \log
      \left(
      \frac
      {1}
      {\kappa \left( d_{ij}^{z}\right)}
      -
      1
      \right)
    \biggl]              = 0
  \end{aligned}
  \label{eq:lim}
\end{equation}

Based on Eq.(\ref{eq:lim}), we find that if $i,j$ is neighbor~($\mathcal{H}_{ij}=1$) and $\alpha\to0$, there is no difference between the CL loss $L_\text{CL}$ and SCL loss $L_{\text{SCL}}$.
When if $\mathcal{H}_{ij}=0$, the difference between the loss functions will be the function of $d_{ij}^{z}$. The CL loss $L_\text{CL}$ only minimizes the distance between adjacent nodes and does not maintain any structural information. The proposed SCL loss considers the knowledge both comes from the output of the current bottleneck and data augmentation, thus less affected by view noise.

\vspace{5mm}

\textbf{Details of Eq.~(\ref{eq_sclcl}).}
Due to the very similar gradient direction, we assume $\dot{Q}_{ij} = 1-Q_{ij}$. The contrastive learning loss is written as, 
\begin{equation}
  \begin{aligned}
    L_\text{CL}  \approx & - \sum
    \left\{
    \mathcal{H}_{ij}
    \log
    Q_{ij}
    +
    \left(1-\mathcal{H}_{ij} \right)
    \log
    \left(1-{Q}_{ij} \right)
    \right\}
  \end{aligned}
\end{equation}
where $\mathcal{H}_{ij}$ indicates whether $i$ and $j$ are augmented from the same original data. 

The SCL loss is written as:


\begin{equation}
  \begin{aligned}
    L_{\text{SCL}} & =
    -
    \sum
    \left\{
    P_{ij}
    \log
    Q_{ij}
    +
    \left(1-P_{ij}\right)
    \log
    \left(1-Q_{ij}\right)
    \right\}
  \end{aligned}
  \label{eq:appendix_SCL}
\end{equation}

According to Eq.~(4) and Eq.~(5), we have

\begin{equation}
  \begin{aligned}
    P_{ij} &= R_{ij} \kappa(d^y_{ij}) = R_{ij} \kappa(y_i, y_j),
       R_{ij} = \left\{
    \begin{array}{lr}
       e^\alpha   \;\;\; \text{if} \;\; \mathcal{H}(x_i, x_j)=1 \\
      1  \;\;\;\;\;\;\;  \text{otherwise}                       \\
    \end{array}
    \right.,                    \\
    Q_{ij} & = \kappa(d_{ij}^z) = \kappa(z_i, z_j),
  \end{aligned}
\end{equation}

For ease of writing, we use distance as the independent variable, $d_{ij}^y=\|y_i- y_j\|_2$, $d_{ij}^z=\|z_i- z_j\|_2$.





The difference between the two loss functions is:


\begin{equation}
  \begin{aligned}
       & L_\text{CL} - L_{\text{SCL}} \\
       =& -\sum\biggl[
    \mathcal{H}_{ij}
    \log \kappa \left( d_{ij}^{z}\right)
    +
    \left(1-\mathcal{H}_{ij} \right)
    \log \left(1-\kappa \left(d_{ij}^{z}\right)\right)
    -
     R_{ij}\kappa\left( d^y_{ij} \right)
    \log \kappa \left( d^z_{ij} \right)
    -
    \left(1- R_{ij}\kappa\left( d^y_{ij} \right)\right)
    \log \left(1-\kappa \left( d^z_{ij} \right)\right)
    \biggl]                   \\
    = & -\sum\biggl[
      \left(
      \mathcal{H}_{ij} -  R_{ij}\kappa\left(d^y_{ij} \right)
      \right)
      \log \kappa \left( d_{ij}^{z}\right)
      +
      \left(
      1-\mathcal{H}_{ij} -1 +  R_{ij}\kappa\left(d^y_{ij} \right)
      \right)
      \log \left(1-\kappa \left(d^z_{ij}\right)\right)
    \biggl]                 \\
    = & -\sum\biggl[
      \left(
      \mathcal{H}_{ij} - R_{ij} \kappa \left(d^y_{ij} \right)
      \right)
      \log \kappa \left( d_{ij}^{z}\right)
      +
      \left(
      R_{ij} \kappa \left(  d^y_{ij} \right) - \mathcal{H}_{ij}
      \right)
      \log \left(1-\kappa \left(d^z_{ij}\right)\right)
    \biggl]                 \\
    = & -\sum\biggl[
      \left(
      \mathcal{H}_{ij} - R_{ij}\kappa \left(d^y_{ij} \right)
      \right)
      \left(
      \log \kappa \left( d_{ij}^{z}\right)
      -
      \log \left(1-\kappa \left(d^z_{ij}\right)\right)
      \right)
    \biggl]                 \\
    = & \sum\biggl[
      \left(
      \mathcal{H}_{ij} - R_{ij} \kappa \left(d^y_{ij} \right)
      \right)
      \log
      \left(
      \frac
      {1}
      {\kappa \left( d_{ij}^{z}\right)}
      -
      1
      \right)
    \biggl]                 \\
  \end{aligned}
  \label{eq:appendix_diff_twoloss_3_0}
\end{equation}

Substituting the relationship between $\mathcal{H}_{ij}$ and $R_{ij}$, $R_{ij} = 1+(e^\alpha -1)\mathcal{H}_{ij}$, we have

\begin{equation}
  \begin{aligned}
    L_{\text{CL}} - L_{\text{SCL}}=\sum
    \biggl[
    \left(
    \mathcal{H}_{ij} - [1+(e^\alpha -1)\mathcal{H}_{ij}] \kappa \left(d^y_{ij} \right)
    \right)
    \log
    \left(
    \frac
    {1}
    {\kappa \left( d_{ij}^{z}\right)}
    -
    1
    \right)
    \biggl] \\
  \end{aligned}
  \label{eq:appendix_diff_twoloss_3_1}
\end{equation}

We assume that network $H(\cdot)$ to be a Lipschitz continuity function, then

\begin{equation}
  \begin{aligned}
    \frac{1}{K} H(d^z_{ij}) \leq d^y_{ij} \leq K H(d^z_{ij}) \quad \forall i, j \in \{1,2,\cdots,N\} 
  \end{aligned}
\end{equation}

We construct the inverse mapping of $H(\cdot)$ to $H^{-1}(\cdot)$,

\begin{equation}
  \begin{aligned}
    \frac{1}{K} d^z_{ij} \leq d^y_{ij} \leq K d^z_{ij} \quad \forall i, j \in \{1,2,\cdots,N\}
  \end{aligned}
\end{equation}

and then there exists $k^*$:
\begin{equation}
  \begin{aligned}
    d^y_{ij} = k^* d^z_{ij} \quad k^* \in [1/K, K] \quad \forall i, j \in \{1,2,\cdots,N\}
  \end{aligned}
  \label{eq:g_revers}
\end{equation}

Substituting the Eq.(\ref{eq:g_revers}) into Eq.(\ref{eq:appendix_diff_twoloss_3_1}).

\begin{equation}
  \begin{aligned}
    L_{\text{CL}} - L_{\text{SCL}}=\sum
    \biggl[
    \left(
    \mathcal{H}_{ij} - [1+(e^\alpha-1)\mathcal{H}_{ij}] \kappa \left(k^* d^z_{ij}\right)
    \right)
    \log
    \left(
    \frac
    {1}
    {\kappa \left( d_{ij}^{z}\right)}
    -
    1
    \right)
    \biggl] \\
  \end{aligned}
  \label{eq:appendix_diff_twoloss_4}
\end{equation}

\clearpage

\subsection{SCL is better than CL in view-noise}

\label{app_proof_3}
To demonstrate that compared to contrastive learning, the proposed SCL Loss has better results, we first define the signal-to-noise ratio~(SNR) as an evaluation metric.
\begin{equation}
  SNR = \frac{PL}{NL}
\end{equation}
where $PL$ means the expectation of positive pair loss, $NL$ means the expectation of noisy pair loss. \\
This metric indicates the noise-robust of the model, and obviously, the bigger this metric is, the better. \\
In order to prove the soft contrastive learning's SNR is larger than contrastive learning's, we should prove:
\begin{equation}
  \frac{PL_{cl}}{NL_{cl}} < \frac{PL_{scl}}{NL_{scl}}\label{prove}
\end{equation}

Obviously, when it is the positive pair case, $S~(z_i, z_j)$ is large if $H~(x_i, x_j)=1$ and small if $H~(x_i, x_j)=0$.
Anyway, when it is the noisy pair case, $S~(z_i, z_j)$ is small if $H~(x_i, x_j)=1$ and large if $H~(x_i, x_j)=0$.\\
First, we organize the $ {PL_{scl} }$ and $ {PL_{cl} }$ into 2 cases,  $H~(x_i, x_j)=1$ and  $H~(x_i, x_j)=0$, for writing convenience, we write $S~(z_i, z_j)$ as $S$ and $S'$, respectively.
\begin{equation}
  P L_{scl}=-M\left\{\left(1-S^{\prime}\right) \log \left(1-S^{\prime}\right)+S^{\prime} \log S^{\prime}\right\}-\left\{\left(1-e^\alpha S\right) \log (1-S)+e^\alpha S \log S\right\}
\end{equation}
\begin{equation}
  P L_{cl}=-M \log \left(1-S^{\prime}\right)-\log S
\end{equation}
M is the ratio of the number of occurrences of $H=1$ to $H=0$.
So, we could get:
\begin{equation}
  \begin{aligned}
     & \ PL_{scl} - PL_{cl}                                                                                                                                                               \\
     & =-M\left\{\left(1-S^{\prime}-1\right) \log \left(1-S^{\prime}\right)+S^{\prime} \log S^{\prime}\right\}-\left\{\left(1-e^\alpha S\right) \log (1-S)+(e^\alpha S -1) \log S\right\} \\
     & =-M\left\{S^{\prime}\left( \log S^{\prime} -\log \left(1-S^{\prime}\right) \right) \right\}-\left\{(e^\alpha S -1) \left(\log S -\log (1-S) \right)\right\}                        \\
     & =-M\left\{S^{\prime} log\frac{ S^{\prime}}{ \left(1-S^{\prime}\right)} \right\}-\left\{(e^\alpha S -1)  log\frac{ S}{ \left(1-S\right)} \right\}
  \end{aligned}
\end{equation}
In the case of positive pair, $S$ converges to 1 and $S^{\prime}$ converges to 0. \\
Because we have bounded that $ e^\alpha S<=1$, so we could easily get:
\begin{equation}
  (e^\alpha S -1)  log\frac{ S}{ \left(1-S\right)} <= 0
\end{equation}
Also, we could get:
\begin{equation}
  -M\left\{S^{\prime} log\frac{ S^{\prime}}{ \left(1-S^{\prime}\right)} \right\} > 0
\end{equation}
So we get:
\begin{equation}
  PL_{scl} - PL_{cl}> 0
\end{equation}
And for the case of noise pair, the values of $S$ and $S'$ are of opposite magnitude, so obviously, there is $NL_{scl} - NL_{cl}< 0$.\\
So the formula Eq.~(\ref{prove}) has been proved.
\clearpage
\section{Details of Balance Hscore}
\label{app_proof_2}

Inspired by the idea of Weighted Harmonic Means, the proposed Balance Hscore is,
\begin{equation}
  \text{Balance Hscore}=B=\frac{1+\theta}{\frac{1}{A_c}+\frac{\theta}{A_t}}=\frac{A_t A_c}{A_t+\theta A_c}(1+\theta)
\end{equation}
where $\theta$ is the ratio of unknown and known samples, 
The $A_c$ is the accuracy of known classes, and $A_t$ is the accuracy of unknown classes.

\textbf{Why Balance Hscore is balance for known classes and unknown classes.} To avoid sacrificing a category's accuracy in exchange for another category's accuracy, we assume that the change in the number of the correct categories and the number of the unknown categories has the same impact on the evaluation metric. 

Let $M$ be the number of the samples of known classes, and $N_c$ be the number of the correct samples of known classes, with $A_c=N_c/M$. The impact of Balance Hscore from the known class is, 
\begin{equation}
  \begin{aligned}
  \frac{\partial B}{\partial N_c} &=\frac{\partial B}{\partial A_c} \cdot \frac{\partial A_c}{\partial N_c} \\
  &=A_t(1+\theta) \cdot \frac{\theta A_c+A_t-\theta A_c}{(\theta A_c+A_t)^2} \cdot \frac{1}{M} \\
  &=\frac{(1+\theta) A_t^2}{M(\theta A_c+A_t)^2} \\
  \end{aligned}
\end{equation}

Let $M_t$ be the number of the samples of known classes, and $N_t$ be the number of the correct samples of known classes, with $A_t=N_t/M_t=N_t/(\theta M)$. The impact of a Balance Hscore from the unknown class is, 
\begin{equation}
  \begin{aligned}
  \frac{\partial B}{\partial N_t} &=\frac{\partial B}{\partial A_t} \cdot \frac{\partial A_t}{\partial N_t} \\
  &=A_c(1+\theta) \cdot \frac{(\theta A_c+A_t)-A_t}{(\theta A_c+A_t)^2} \cdot \frac{1}{\theta M} 
  =\frac{(1+\theta) A_c^2}{M(\theta A_c+A t)^2} \\
  \end{aligned}
\end{equation}

So if $ A_c = A_t $, we have $$\frac{\partial B}{\partial N_c}=  \frac{\partial B}{\partial N_t},$$ 
it indicates that the metric gets the same influence as the correct classification. Thus the Balance Hscore is balance for known and unknown classes.

\textbf{Why Hscore is unbalance for known classes and unknown classes.}
However, for the 
$$\text{Hscore} = (2\cdot A_c \cdot A_t) / (A_c +A_t).$$

The impact of the Hscore by the known class is
\begin{equation}
  \begin{aligned}
  \frac{\partial H}{\partial N_c} &=\frac{\partial H}{\partial A_c} \cdot \frac{\partial A_c}{\partial N_c} \\
  &=2 A_t \cdot \frac{A_t+A_c-A_c}{(A_c + A_t)^2} \cdot \frac{1}{M} \\
  &=\frac{2 A_t^2}{M (A_c + A_t)^2}
  \end{aligned}
\end{equation}

The impact of the Hscore by the unknown class is
\begin{equation}
  \begin{aligned}
  \frac{\partial H}{\partial N_t} &=\frac{\partial H}{\partial A_t} \cdot \frac{\partial A_t}{\partial N_t} \\
  &=2 A_c \cdot \frac{A_c+A_t-A_t}{(A_c+A_t)^2} \cdot \frac{1}{\theta M} \\
  &=\frac{2 A_c^2}{\theta M(A_c t A_t)^2} \\
  \end{aligned}
\end{equation}
So when $ A_c = A_t $, we could get$ \frac{\partial B}{\partial N_c} \neq \frac{\partial B}{\partial N_t}$, we think it's not balance.\\

\clearpage
\section{Experimental setups}

\subsection{Baseline Methods}

We aim to compare methods of universal domain adaptation~(UNDA), which can reject unknown samples, such as, CMU~\cite{fu2020learning}, DANCE~\cite{saito2020universal}, DCC~\cite{GuangruiLi2021DomainCC},  OVANet~\cite{KuniakiSaito2021OVANetON}, TNT~\cite{LiangChen2022EvidentialNC}, GATE~\cite{LiangChen2022GeometricAC} and D+SPA~\cite{kundu2022subsidiary}. 
We are looking at some contemporaneous work such as KUADA~\cite{YifanWang2022ExploitingIA}, UACP~\cite{YunyunWang2022TowardsAU} and UEPS~\cite{YifanWang2022ANF}, which we did not include in the comparison because the source code was not available and some of these works were not peer-reviewed. Instead of reproducing the results of these papers, we directly used the results reported in the papers with the same configuration. 

\subsection{Datasets}

We utilize popular datasets in DA: Office~\cite{saenko2010}, OfficeHome~\cite{venkateswara2017Deep}, VisDA~\cite{peng2017visda}, and DomainNet~\cite{peng2018moment}. Unless otherwise noted, we follow existing protocols~\cite{KuniakiSaito2021OVANetON} to split the datasets into source-private ($|L_s-L_t|$), target-private ($|L_t-L_s|$) and shared categories ($|L_s \cap L_t|$).

\begin{table}[h]
  \centering
  \begin{center}
    \begin{minipage}{0.6\textwidth}
      \caption{The division on label sets in each setting}\label{tab1}%
      \begin{tabular}{llccc}
        \hline Tasks & Datasets & \(\left|L_{s} \cap L_{t}\right|\) & \(\left|L_{s}-L_{t}\right|\) & \(\left|L_{t}-L_{s}\right|\) \\
         \hline \multirow{2}{*}{ ODA } & Office-31 & 10 & 0 & 11 \\ 
         & VisDA & 6 & 0 & 6 \\ \hline \multirow{4}{*}{ UNDA } & Office-31 & 10 & 10 & 11 \\ 
         & Office-Home & 10 & 5 & 50 \\ & VisDA & 6 & 3 & 3 \\ 
         & DomainNet & 150 & 50 & 145 \\ 
         \hline
      \end{tabular}
    \end{minipage}
  \end{center}
\end{table}

\subsection{Top\_n softmax in AIO}
\label{app_Top_nsoftmax}

The forward propagation of $C^\text{AIO}\left( \cdot \right)$ is
\begin{equation}
  \mathcal{C}_{x_i} = \left\{c_{x_i}^k,\tilde{c}_{x_i}^k| k\in \mathcal{K} \right\} = \sigma\left( C^\text{AIO}\left(z_{x_i}\right)\right),
\end{equation}
The $c_{x_i}^k$ and $\tilde{c}_{x_i}^k$ are the probability of $x_i$ being identified as a known and unknown class by $k$th category, $\sum_{k} \left\{c_{x_i}^k + \tilde{c}_{x_i}^k\right\}=1$.
The $\sigma(\cdot)$ is a `top\_n softmax' function to ensure $ \sum_{k\in \mathcal{T}^N} \{c_{x_i}^k+\tilde{c}_{x_i}^k\}=1$, $\mathcal{T}^N$ is the top $N=20$ item of $\mathcal{C}_{x_i}$.
We deploy `top\_n softmax' to balance the loss scale of different category numbers. For example, in UNDA setting, there are 200 known categories in DomainNet, while only 20 known categories in Office. If deploying a simple softmax, the loss scale will vary over a wide range with different datasets.
\clearpage

\end{document}